\documentclass[format=sigconf, 9pt]{acmart}

\usepackage{booktabs}

\usepackage{multirow}

\usepackage[labelsep=period]{caption}

\usepackage[]{graphicx}

\usepackage{tikz}

\usepackage[export]{adjustbox}
\usepackage[font=small]{subfig}

\usepackage{amsmath}
\usepackage{amsfonts}

\usepackage{amsthm}

\usepackage[T1]{fontenc}

\usepackage{balance}

\usepackage{siunitx}

\usepackage{enumitem}

\usepackage{soul}

\usepackage{xcolor}

\usepackage{algorithm}
\usepackage{algpseudocode}

\setcopyright{none}

\begin{document}
	
\renewcommand\footnotetextcopyrightpermission[1]{} 
\pagestyle{plain} 

	\title{NullaNet: Training Deep Neural Networks for Reduced-Memory-Access Inference}

\author{Mahdi Nazemi}
\affiliation{%
	\institution{University of Southern California}
	\city{Los Angeles}
	\state{California}
	\country{USA}
}
\email{mnazemi@usc.edu}

\author{Ghasem Pasandi}
\affiliation{%
	\institution{University of Southern California}
	\city{Los Angeles}
	\state{California}
	\country{USA}
}
\email{pasandi@usc.edu}

\author{Massoud Pedram}
\affiliation{%
	\institution{University of Southern California}
	\city{Los Angeles}
	\state{California}
	\country{USA}
}
\email{pedram@usc.edu}
		
\begin{abstract}
	Deep neural networks have been successfully deployed in a wide variety of applications including computer vision and speech recognition. 
	However, computational and storage complexity of these models has forced the majority of computations to be performed on high-end computing platforms or on the cloud. 
	To cope with computational and storage complexity of these models, this paper presents a training method that enables a radically different approach for realization of deep neural networks through Boolean logic minimization. 
	The aforementioned realization completely removes the energy-hungry step of accessing memory for obtaining model parameters, consumes about two orders of magnitude fewer computing resources compared to realizations that use floating-point operations, and has a substantially lower latency. 
\end{abstract}

\maketitle

\section{Introduction}

	Deep neural networks (DNNs) have proven successful in a wide variety of applications such as speech recognition and synthesis, computer vision, machine translation, and game playing, to name but a few. 
	To sustain the ubiquitous deployment of deep learning models and cope with their computational and storage complexity, several solutions like weight pruning, knowledge distillation, tensor decomposition, and quantization have been proposed. 
	The ultimate goal of these methods is to reduce the cost of computations and/or memory accesses without affecting classification accuracy significantly. 

	To demonstrate the cost associated with computations and memory accesses, Table~\ref{table:intel-delay} summarizes latency values for integer arithmetic operations and accesses to different levels of the memory hierarchy in Intel Haswell architecture \cite{intel-opt-manual, intel-haswell}. 
	The latency values for Skylake architecture are close to the ones reported in Table~\ref{table:intel-delay} \cite{intel-skylake}. 
	It can be observed that accessing different levels of memory hierarchy is $4\times$ to $400\times$ slower than integer operations. 
	This slow operation of memories is a major bottleneck for deep neural networks not only because they have millions of parameters that need to be read from memory, but also  because of frequent reading and writing of intermediate values of arithmetic operations.

	\begin{table}[b]
		\centering
		\captionsetup{justification=centering}
		\caption{Latency values for 32-bit integer operations and memory accesses in Intel Haswell architecture}
		\label{table:intel-delay}
		\resizebox{0.9\columnwidth}{!}{%
			\begin{tabular}{l cc}
				\textbf{Integer Operation}				& \textbf{\# of Operations}					& \textbf{Latency} (Clock Cycle)	\\
				\midrule
				Add										& 12             							& 1									\\
				Multiply								& 4             							& 1									\\			
				\midrule
				\midrule
				\textbf{Memory}							& \textbf{Size} (KBytes)					& \textbf{Latency} (Clock Cycle)	\\
				\midrule
				L1 Data Cache							& 32										&  4 -- 5							\\
				L2 Cache								& 256										&  12								\\
				L3 Cache								& 8192										&  36 -- 58							\\
				DRAM									& --										&  230 -- 422															
			\end{tabular}
		}
	\end{table}

	Table~\ref{table:energy-45nm} compares different types of arithmetic operations and memory accesses from an energy consumption point of view \cite{horowitz20141}. 
	It can be observed that energy consumption of accessing level-1 cache is about $5\times$ that of half-precision floating-point multiplication. 
	Moreover, accessing DRAM consumes about $300\times$ -- $600\times$ more energy compared to half-precision floating-point multiplication. 
	This high energy consumption of memory elements necessitates designing memory-efficient methods for implementing deep neural networks, especially for energy-constrained platforms such as smartphones. 

	\begin{table}[b]
		\centering
		\captionsetup{justification=centering}
		\caption{Energy cost of various arithmetic operations and memory accesses in 45nm technology}
		\label{table:energy-45nm}
		\resizebox{0.9\columnwidth}{!}{%
			\begin{tabular}{l ccc}
				\textbf{Arithmetic}				& \multirow{2}{*}{\textbf{Bit Width}}			& \textbf{Energy}			& {\textbf{Normalized Energy}}		\\
				\textbf{Operation}				& {}											& (\si{\pico\joule})			& (per bit)				\\
				\midrule
				Integer Add						& 32    										& 0.1 						& 1									\\
				Integer Multiply				& 32            								& 3.1 						& 31								\\
				Float Add						& 16            								& 0.4 						& 8									\\
				Float Add						& 32            								& 0.9 						& 9									\\
				Float Multiply					& 16 											& 1.1 						&	22								\\
				Float Multiply					& 32            								& 3.7 						&	37								\\	
				\midrule
				\midrule
				\multirow{2}{*}{\textbf{Memory}}& \multirow{2}{*}{\textbf{Bit Width}}			& \textbf{Energy}			& {\textbf{Normalized Energy}}		\\
				{}								& {}											& (\si{\pico\joule})		& (per bit)				\\
				\midrule
				L1 Data Cache					& 64											&  20						& 100								\\
				DRAM							& 64											&  1,300 -- 2,600			& 6,500 -- 13,000
			\end{tabular}
		}
	\end{table}

	This paper presents a method for realizing deep neural networks that requires no memory accesses for reading model parameters. 
	In this method, different layers of a deep neural network are trained to have binary input/output activations and floating-point weights. 
	Layers that are trained this way can be treated as multi-input multi-output Boolean functions, which can be optimized and implemented efficiently using Boolean logic optimization algorithms. 
	One of the major advantages of the present work compared to binarized neural networks is that it does not quantize weights and biases to any specific values, and therefore, does not affect the classification accuracy as much. 
	In fact, similar to regular deep neural networks, weights and biases can assume any value. 
	Such realization of neural networks reduces computational cost, power consumption, and latency by a large margin compared to floating-point or fixed-point implementations. 
	
	The remainder of this paper is organized as follows. 
	Section~\ref{sec:related-work} reviews the related work and details the shortcomings of some of the prior work compared to our proposed method. 
	Section~\ref{sec:methodology} describes methodology.
	After that, Section~\ref{sec:results} presents experimental results and finally, Section~\ref{sec:conclusion} concludes the paper. 

\section{Related Work} \label{sec:related-work}

	There has been a considerable amount of work on ameliorating computational and/or memory cost of deep neural networks. 
    The majority of work that tackle computational and/or memory cost of DNNs can be classified into two categories: 
    the ones that introduce new methods for training DNNs such that they can be realized more efficiently, and those that accelerate networks trained using existing methods by introducing a paradigm shift in how computations are performed.
    Weight pruning, knowledge distillation, tensor decomposition, and quantization are among successful methods for improving computational and/or memory cost of realizing DNNs. 

	The idea of weight pruning is to remove weights of a DNN that are below a threshold, and optionally, retrain the reduced network to compensate for accuracy degradation \cite{han2015learning}.
    This process can be repeated a few times to find a more compact DNN. 
    The compact DNN has fewer parameters compared to the original network, and therefore, requires fewer computations and less storage. 
    Han \textit{et~al.} \cite{han2016eie} designed a specialized hardware that works directly on these compact DNNs and were able to achieve speedup and energy savings compared to floating-point-based implementations of regular deep neural networks.

    Knowledge distillation methods take a different approach for finding networks that reach an acceptable classification accuracy while they require reasonably fewer computations and accesses to memory \cite{hinton2015distilling}.
    In these methods, a large, complex network (teacher) imparts its knowledge to a simpler network (student). 
    This allows the simple network to achieve a classification accuracy that would be unachievable if it was trained directly on the same dataset. 

    Quantization methods use representations that are more efficient in terms of computation and/or storage compared to single-precision floating-point representation. 
    This includes representing model parameters, activations, and/or gradients with a different data type such as fixed-point, or a representation that consumes fewer number of bits (e.g. fewer number of bits for representing exponent and mantissa of floating-point representation). 
    Examples of using other data types and/or low-bit-width representations can be found in \cite{li2016ternary,venkatesh2017accelerating,miyashita2016convolutional,zhou2016dorefa,koster2017flexpoint,nazemi2018deploying,ghasemzadehrebnet,chung2018serving,polino2018model,achterhold2018variational,hou2018loss,khoram2018adaptive}. 

    An extreme case of quantization is representing each weight and activation with a binary value (i.e., a single bit). 
    Hubara \textit{et~al.} \cite{hubara2016binarized} introduce binarized neural networks where weights and activations take binary values (i.e. $-1$ and $+1$). 
    This type of quantization allows replacing multiplication with a simple XNOR operation (XNOR($-1,-1$) = XNOR($1,1$) = $1$; and XNOR($-1,1$) = XNOR($1,-1$) = -1). 
    On the surface, representing each value (i.e. weights and activations) with a single bit achieves $32\times$ memory cost reduction compared to single-precision floating-point representation. 
    However, as we will explain shortly, binarized neural networks require a higher storage for saving model parameters compared to networks that use floating-point representation. 

    Representing each value with a single bit incurs a high degradation in classification accuracy due to the quantization error. 
    As a result, wider or larger binarized neural networks need to be trained to achieve the same classification accuracy as networks that utilize floating-point values and operations. 
    For example, a binarized neural network with three hidden layers and 4,096 neurons per layer \cite{hubara2016binarized} achieves 99.04\% classification accuracy on the MNIST dataset \cite{lecun2010mnist}. 
    The number of trainable parameters in this network is about 36.80 millions and \SI{4.39}{MB} storage is required for saving these parameters for inference. 
    On the other hand, a neural network that uses floating-point representation needs two hidden layers with 512 neurons per layer \cite{tang2013deep} to achieve a 99.13\% accuracy on the same dataset. 
    The number of trainable parameters in the latter network is about 303.10 thousands, which is equivalent to \SI{1.03}{MB} storage requirement for saving model parameters. 
    It can be observed that storage requirement for the binarized network is about $3.8\times$ that of the neural network that uses floating-point values,  and that this ratio will be doubled if half-precision floating-point representation is used in the latter network. 
    The number of computations in the binarized network is about $121\times$ that of the latter network, however, each computation is an XNOR operation instead of a floating-point multiplication. 

    XNOR-Net \cite{rastegari2016xnor} reduces quantization error of binarized neural networks by effectively implementing $-w$ and $+w$ weights instead of $-1$ and $+1$ weights. 
    On the other hand, it incurs additional costs in computation, storage, and latency because of the newly introduced floating-point multiplications. 
    One of the major advantages of the present work compared to binarized neural networks and XNOR-Net is that it does not quantize weights and biases to any specific values, and therefore, the classification accuracy is not affected as much. 
    In fact, similar to regular deep neural networks, weights and biases can assume any value. 
    There has been a few attempts at designing specialized hardware for realizing deep neural networks based on binarized neural networks or XNOR-Net. 
    Examples of such implementations can be found in \cite{andri2016yodann,bahou2018xnorbin,umuroglu2017finn}. 

\section{Methodology} \label{sec:methodology}

	This section explains details of a training method which allows the transformation of the DNN realization problem to a Boolean logic optimization problem. 
    Furthermore, it presents how Boolean logic optimization can be applied to a given neural network in order to find an efficient realization in terms of computational resource consumption, memory cost, and latency. 

\subsection{Training}

	The specific goal of the training process in this work is to limit activations to binary values, so that each layer can be modeled as a multi-input multi-output Boolean function. 
    To confine activations to binary values, one can apply the \textit{sign} function as the activation function of a layer. 
    However, the fact that the derivative of the sign function is zero (almost everywhere) prevents the back-propagation algorithm from properly updating model parameters. 
	Bengio \textit{et~al.} \cite{bengio2013estimating} have studied estimation of gradients for neurons with hard non-linearities and have shown that a straight-through estimator (STE) \cite{hinton2012lecture} achieves the best validation and test error. 
    This work uses the same STE that was presented in \cite{hubara2016binarized} and propagates gradients through a \textit{hard tanh} function 
	\begin{equation*}
		Htanh(x) = max(-1, min(1, x)).
	\end{equation*}

    Algorithm~\ref{alg:forward} summarizes different steps of forward propagation for deep neural networks with binary activations \footnote{When discussing weights, biases, and/or activations in this paper, subscripts are used for indexing layers while superscripts are used for indexing neurons.} (steps that are limited to specific types of layers, e.g., max pooling, are not shown). 
    Derivatives of all operators (used in back-propagation algorithm) are as usual, except for the sign function which is estimated with the aforementioned STE. 
    These paradigms for forward and backward propagations are used to train neural networks. 
    In a network that is trained this way, the first layer has decimal inputs and binary outputs, the last layer has binary inputs and decimal outputs, and all other layers have binary inputs and binary outputs. 
    Layers with binary inputs and outputs are the ones that can be optimized using Boolean logic minimization techniques. 
    The first and last layers can be optimized differently, which will be explained later in the paper. 

    \begin{algorithm}[t]
    	\caption{Forward Propagation}
        \label{alg:forward}
    		
		\begin{algorithmic}[1]
    		\Require
            	\Statex $L$: \text{number of layers}
    			\Statex $\mathbf{a}_0$: \text{a mini-batch of inputs}
    			\Statex $\mathbf{W}$: \text{weights}
    			\Statex $\boldsymbol{\beta}$: \text{batch normalization parameters}
   			\Ensure
   				\Statex $\mathbf{a}_L$: \text{network's predictions}
    			
    		\For {$i = 1$ to $L$}
    			\State $\mathbf{z}_i = \mathbf{a}_{i - 1} \mathbf{W}_{i}$
                \State $\mathbf{a}_i = BatchNorm(\mathbf{z}_i, \boldsymbol{\beta})$
				\If {$i < L$}
    				\State $\mathbf{a}_i = Sign(\mathbf{a}_i)$
    			\EndIf
    		\EndFor
    			
   			\State return $\mathbf{a}_L$
    	\end{algorithmic}
	\end{algorithm}

\subsection{Boolean Logic Optimization}

\begin{figure}[t]
            \centering
            \subfloat[] {
				\includegraphics[width=0.3\columnwidth, valign=c]{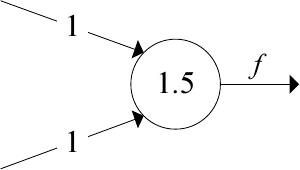}
				\label{fig:AND}
			}
            \subfloat[] {
				\includegraphics[width=0.3\columnwidth, valign=c]{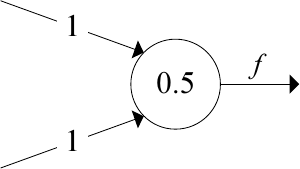}
				\label{fig:OR}
			}
            \subfloat[] {
				\includegraphics[width=0.3\columnwidth, valign=c]{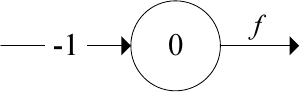}
				\label{fig:NOT}
			}
            
            \subfloat[] {
				\includegraphics[width=0.5\columnwidth, valign=c]{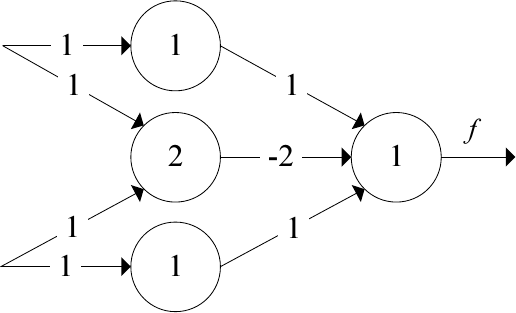}
				\label{fig:XOR}
			}
            
            \caption{Examples of implementing logic gates using McCulloch-Pitts neurons: (a)~AND, (b)~OR, (c)~NOT, and (d)~XOR. Weights are shown on edges while biases are shown inside circles.}
            \label{fig:mp-gates}
    \end{figure}

	During early developments of neural networks, one of the initial problems was to implement AND, OR, and NOT gates using neurons. 
    The rationale behind this was that because computers can be built using these three gates, implementing them using neurons allows building computers based on neural networks \cite{arbib2012brains}. 
    Fig.~\ref{fig:mp-gates} illustrates examples of logic gates implemented using \textit{McCulloch-Pitts} neurons. 
	McCulloch-Pitts neurons are defined according to Eq.~\ref{eq:mp-neuron}  
	\begin{equation}
  		\label{eq:mp-neuron}
  		f = 
        \begin{cases}
        	1, & \text{if } \sum\limits_{j}^{} a^j \times w^j \geq b \\
			0, & \text{otherwise}
		\end{cases}
	\end{equation}
    where $a^j$ and $w^j$ denote the \textit{j}-th input and weight of the neuron, respectively, and $b$ is the neuron's bias (or threshold). 
   	This section explains a few methods that are capable of doing the reverse operation: taking a neuron as an input and implementing its functionality using primitive gates such as AND, OR, NOT, and XOR.

\subsubsection{Realization Based on Input Enumeration} \label{sec:enumeration}

	One of the methods that allows inference without storing model parameters is realization based on input enumeration. 
    In this method, all different combinations of a neuron's inputs are enumerated and the corresponding outputs are found according to the neuron's weights and bias (Eq.~\ref{eq:mp-neuron}). 
    This is in fact equivalent to finding a truth table for each neuron. 
    The truth table implements the same function as the neuron's function when its output is calculated using Eq.~\ref{eq:mp-neuron}. 

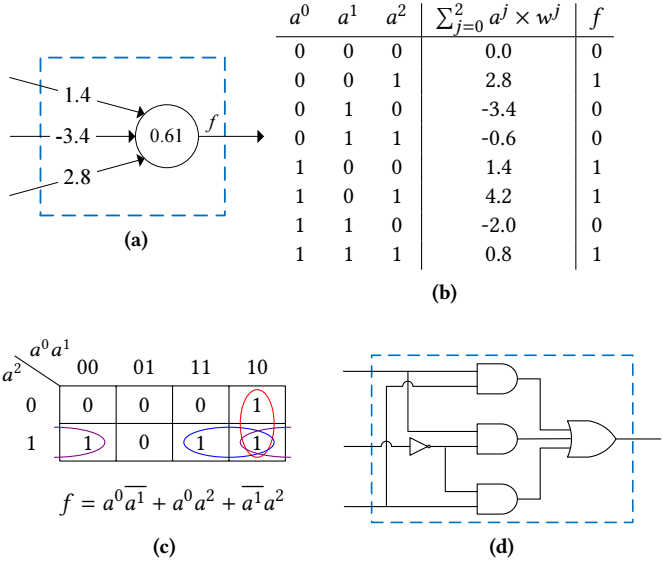
\begin{figure}[tb]
            \centering
            \subfloat[] {
				\includegraphics[width=0.4\columnwidth, valign=c]{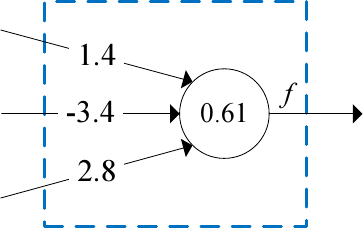}
			}
            \subfloat[] {
            	\begin{tabular*}{0.53\columnwidth}{ccc | c | c}
                	$a^0$		& $a^1$		& $a^2$		& $\sum_{j = 0}^{2} a^j \times w^j$		& $f$		\\
                	\hline
                	0			& 0			& 0			& 0.0									& 0			\\
                    0			& 0			& 1			& 2.8									& 1			\\
                    0			& 1			& 0			& -3.4									& 0			\\
                    0			& 1			& 1			& -0.6									& 0			\\
                    1			& 0			& 0			& 1.4									& 1			\\
                    1			& 0			& 1			& 4.2									& 1			\\
                    1			& 1			& 0			& -2.0									& 0			\\
                    1			& 1			& 1			& 0.8									& 1			\\
            	\end{tabular*}
            }
            
            \subfloat[] {
				\begin{tikzpicture}[x=.75cm,y=.5cm]
                  \draw (1,0) grid [step=1] (5,2);

                  \draw (0.2,2.8) -- (1,2);
                  \node at (0.4,2.7) [below left,inner sep=1pt] {\small$a^2$};
                  \node at (0.4,2.7) [above right,inner sep=1pt] {\small$a^0 a^1$};

                  \node at (1.5,2.5) {00};
                  \node at (2.5,2.5) {01};
                  \node at (3.5,2.5) {11};
                  \node at (4.5,2.5) {10};

                  \node at (0.5,0.5) {1};
                  \node at (0.5,1.5) {0};

                  \node at (1.5,1.5) {0};
                  \node at (2.5,1.5) {0};
                  \node at (3.5,1.5) {0};
                  \node at (4.5,1.5) {1};

                  \node at (1.5,0.5) {1};
                  \node at (2.5,0.5) {0};
                  \node at (3.5,0.5) {1};
                  \node at (4.5,0.5) {1};

                  \draw [red] (4.8,1) arc[start angle=0,end angle=360, x radius=0.3, y radius=0.9];

                  \draw [blue] (4.8,0.5) arc[start angle=0,end angle=360, x radius=0.8, y radius=0.4];

                  \draw [violet] (5.1,0.9) arc[start angle=90,end angle=270, x radius=0.9, y radius=0.4];
                  \draw [violet] (0.9,0.1) arc[start angle=-90,end angle=90, x radius=0.9, y radius=0.4];
                  
                  \node [below=0.2cm, align=flush center,text width=4cm] at (3.0,0.0)
                  	{
						$f = a^0 \overline{a^1} + a^0 a^2 + \overline{a^1} a^2$
					};

				\end{tikzpicture}
			}
            \subfloat[] {
				\includegraphics[width=0.5\columnwidth]{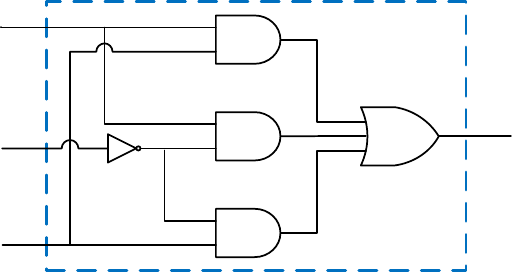}
			}
            
            \caption{Efficient realization of a neuron using logic gates through finding the neuron's truth table followed by Karnaugh map simplification.}
            \label{fig:mac-to-gates}
    \end{figure}

    Given the truth table, one can write the function of each neuron as a sum of product terms (SoP). 
    The SoP for each neuron can be fed to a logic synthesis tool \cite{sentovich1992sis,mishchenko2007abc}, which implements the function using logic gates and optimizes the function for minimum area, delay, and/or power consumption. 
    In other words, instead of realizing the output of a neuron through calculating the dot product of its inputs and weights, we implement the output using logic gates thorough synthesizing its Boolean expression.  
    Fig.~\ref{fig:mac-to-gates} illustrates an example of such realization. 
    The advantage of such implementation is that the function represented using an SoP considers the neuron's parameters implicitly and allows inference without reading model parameters from memory. 

\begin{figure}[tb]
            \centering
            \includegraphics[width=\columnwidth]{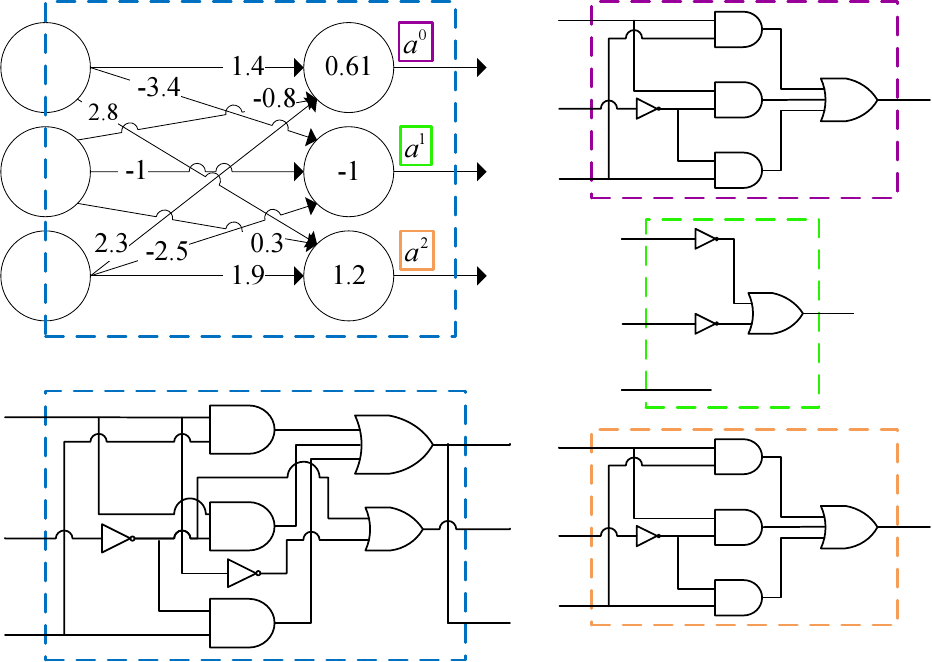}
            \caption{Optimizing neurons in a layer by extracting common logic. The optimized implementation requires seven logic gates while implementing neurons individually will require 13 logic gates.}
            \label{fig:group}
    \end{figure}

    After optimizing all neurons, the neurons of each layer are put together (and optionally, further optimized as a larger block) to find an efficient realization for each layer. 
    Fig.~\ref{fig:group} illustrates an example of such optimization. 
    Similarly, the optimized layers are put together to find an efficient realization for the whole network. 

    The disadvantage of this method is that it can only be applied to neurons with limited number of inputs because the size of truth table grows exponentially with the cardinality of inputs to a neuron. 
    Depthwise separable convolutions are examples of layers that can be implemented using this method. 
    In practice, this solution will be infeasible for neurons with tens or hundreds of inputs. 
    The method explained in the next section addresses this issue. 

\subsubsection{Realization Based on Incompletely Specified Functions (ISFs)}

	An incompletely specified function is a Boolean function where output values are defined only for a subset of input combinations. 
    The input combinations that cause a logic one in the output constitute the ON-set and the input combinations that cause a logic zero in the output constitute the OFF-set. 
    The input combinations for which the output value is not specified make up the DON'T CARE-set (or DC-set for short). 

	In this method, instead of enumerating all input combinations for each neuron, we only evaluate outputs of neurons for input combinations derived from samples in the training set and add the remaining input combinations to the DC-set of the neuron. 
    As a result, the cardinality of ON-set and OFF-set will be linear functions of the cardinality of the training set, rather than an exponential function of the number of neurons' inputs.

    Realizing DNNs based on ISFs has a few advantages. 
    Similar to the method explained earlier, this technique allows inference without storing model parameters explicitly. 
	In other words, the realized logic of each ISF considers the neuron's parameters implicitly and does not require accessing memory for reading the neuron's weights and bias. 
    This results in substantial savings in latency and energy consumption according to Table~\ref{table:intel-delay} and Table~\ref{table:energy-45nm}. 
    Furthermore, as we will explain later, presence of the DC-set allows optimizing logic to a great degree, which translates into considerably lower resource consumption and substantially lower latency compared to using multiplier--accumulators (MACs). 
    There are other advantages to realization based on ISFs which will be explained shortly. 

	Algorithm~\ref{alg:optimization} presents a hierarchical optimization method for optimizing realization of deep neural networks that are trained based on Algorithm~\ref{alg:forward}. 
    In this algorithm, it is assumed that all training data is applied to the trained network as a single batch and activations at different layers are found and provided as one of the inputs to the algorithm. 
    The next few paragraphs explain the details of each step of the algorithm. 

    \begin{algorithm}[b]
    	\caption{Deep Neural Network Optimization}
        \label{alg:optimization}
    		
		\begin{algorithmic}[1]
    		\Require
            	\Statex $L$: \text{number of layers}
                \Statex $u_i, \quad i = 0, 1, 2, ..., L$: \text{number of neurons in layer $i$}
    			\Statex $a_i, \quad i = 0, 1, 2, ..., L$: \text{activations at layer $i$}
                \Statex $\qquad \qquad \qquad \qquad$ \text{(for all training samples)}
   			\Ensure
   				\Statex \text{optimized network}
    			
    		\For {$i = 2$ to $L - 1$}
                \For {$j = 0$ to $u_i - 1$}
                	\State $OptimizeNeuron(Inputs(i, j), Outputs(i, j))$
            	\EndFor
                \State $OptimizeLayer()$
                \State $Pythonize()$
    		\EndFor
            \State $network = OptimizeNetwork()$
    			
   			\State return $network$
    	\end{algorithmic}
	\end{algorithm}

    $OptimizeNeuron(.)$ is a function that takes ISF representation of each neuron and finds a minimal representation in disjunctive normal form for covering the neuron's ON-set. 
    The objective of this step of the optimization is to take advantage of DC-set in finding a cover for the ON-set that has the fewest possible number of cubes (i.e. conjunctive clauses) and fewest possible number of literals per cube. 
    Because the output of an ISF for DC-set is not specified, it can take either the logic zero or logic one during optimization. 
    Typically, the elements of DC-set that are close to elements of ON-set in the \textit{n}-dimensional input space are assigned a value of one and the ones that are close to elements of OFF-set are assigned a value of zero. 
    This is particularly useful in realization of deep neural networks because the input combinations that are not encountered during applying training data to the network (i.e. $a_i$ inputs in Algorithm~\ref{alg:optimization}) will have the same output as the ones that are previously encountered and are close to them. 
    This step of optimization is similar to two-level logic minimization, which is known to be a hard problem and different heuristics have been developed for solving it \cite{brayton1984logic,coudert2002two,fivser2004two}. 
    This step is, in fact, similar to what was illustrated in Fig~\ref{fig:mac-to-gates}, but takes DC-set into account. 

    $OptimizeLayer()$ is the next optimization step which applies combinational logic synthesis algorithms to all neurons that constitute a layer. 
    Because different neurons of a layer share the same inputs, combinational synthesis algorithms may be able to find equivalent logic expressions that are used in different neurons.
    This allows common logic expression extraction (similar to Fig.~\ref{fig:group}) i.e., implementing the shared logic once and providing its output to all corresponding neurons instead of implementing the logic separately for each neuron.
    ABC \cite{mishchenko2007abc} uses algorithms such as rewriting \cite{mishchenko2006dag}, balancing, and refactoring \cite{brayton1982decomposition} to optimize combinational designs. 
    Combinational synthesis algorithms are capable of performing a variety of other advanced optimizations \cite{hachtel2006logic} that are not discussed here. 

    $Pythonize()$ is a step that converts optimized Boolean expression of a layer into Python code that can be run on CPUs and GPUs. 
    This not only allows studying the effect of using ISFs on classification accuracy, but also enables more efficient inference on the aforementioned platforms. 
    Because the output of this step is a layer implemented using primitive gates such as AND, OR, NOT, and XOR, the layer can be implemented efficiently on different computing platforms. 

    $OptimizeNetwork()$ is the last (optional) step of optimization that implements pipelining to increase the throughput of the design. 
    Because each optimized layer is realized using a combinational logic, realization of the whole network will have a large combinational delay. 
    Pipelining is a well-known method that breaks a large combinational design into smaller parts with lower delays. 
    One may use both \textit{macro-pipelining} and \textit{micro-pipelining} to increase the throughput. 
    A stage of the macro-pipeline includes a group of consecutive layers of the network while a stage in the micro-pipeline includes necessary logic for implementing those layers. 
    Combinational synthesis algorithms are applied to layers that lie within a macro-pipeline stage in order to further optimize those layers across their boundaries (i.e. inputs and outputs). 

    After optimizing layers with binary input and output activations using Algorithm~\ref{alg:optimization}, the first and last layers need to be optimized to achieve an overall acceptable performance.  
    One way of optimizing these two layers is to use fixed-point quantization to improve resource consumption and energy-efficiency. 
    A second method of optimization, which can be used in conjunction with the first method, is to store the parameters associated with these layers in a low-latency, low-power memory such as L1 data cache. 
    Because these layers are the only ones that need to access weights and biases, small, low-cost memories can store their parameters efficiently. 

    We should note that the amount of computations in the first layer is typically much smaller than the rest of layers in a \textit{deep} neural network and therefore, the performance of this layer will have little impact on the network's overall performance.
    It should also be noted that for the last layer, because all inputs are pseudo-Boolean variables, dot product of inputs and weights is replaced with additions and subtractions, which are more efficient operations than MACs used in dot product calculation. 

\section{Experimental Results} \label{sec:results}

\subsection{Experimental Setup}
\subsubsection{Dataset} \label{sec:dataset}

	In order to demonstrate the applicability of the proposed method in practice, we optimize two different neural networks for the MNIST dataset of handwritten digits \cite{lecun2010mnist}. 
	This dataset includes 60,000 samples for training and 10,000 samples for testing, where each sample is a 28$\times$28 grayscale image. 
    The last 10,000 samples of the training set are used as validation set for model selection. 
	The objective is to classify each image into one of ten classes 0--9. 

\subsubsection{Training Procedure} \label{sec:train-procedure}

	All networks presented in this section are trained for 100 epochs, with a mini-batch size of 64, and with dropout. 
    For all networks, negative log likelihood loss is minimized using Adamax \cite{kingma2014adam} optimization method. 
    The learning rate is initially set to 0.003 and is gradually decreased during training. 

\subsubsection{Hardware Setup} \label{sec:train-procedure}

	The target platform is an Intel Arria 10 GT 1150 FPGA, which includes 427,200 adaptive logic modules (ALMs), 55,562,240 bits of block RAM, and 1,518 DSP blocks. 
	Table~\ref{table:mac-alm} summarizes resource utilization, clock frequency, latency, and power consumption for implementing single-precision (i.e. 32-bit) and half-precision (i.e. 16-bit) floating-point adders, multipliers, and unfused MACs on the target FPGA. 
    All reported values are found after placement and routing. 
    The Verilog designs for these operations are based on \cite{zhemao-chisel-float}. 
    Both adders and multipliers are pipelined and have four and two pipeline stages, respectively. 
    MACs are implemented using these adders and multipliers and have six pipeline stages. 
    The adders, multipliers, and MACs are realized using ALMs instead of DSPs to allow comparison of our work with designs that are based on floating-point operations. 

	\begin{table}[h]
		\centering
		\captionsetup{justification=centering}
		\caption{Hardware realization results for floating-point adders, multipliers, and unfused MACs}
		\label{table:mac-alm}
		\resizebox{1\columnwidth}{!}{%
			\begin{tabular}{l ccccc}
				\textbf{Arithmetic}		& \multirow{2}{*}{\textbf{ALMs}}	& \textbf{Registers}	& \textbf{Clock Frequency}			& \textbf{Latency}			& \textbf{Power}				\\
				\textbf{Operation}		& {}								& (bits)				& (\si{\MHz})						&  (\si{\ns})				 & (\si{\mW})					\\
				\midrule
				Add (16)				& 115								& 120					& 393.08							& 10.18						& 66.44									\\
				Multiply (16)			& 86								& 56					& 263.85							& 7.58						& 57.79									\\
				MAC (16)				& 195								& 191					& 281.37							& 21.32						& 68.18									\\
                \midrule
				Add (32)				& 253								& 247					& 295.77							& 13.52						& 81.05									\\
				Multiply (32)			& 302								& 101					& 181.00							& 11.05						& 80.77									\\
				MAC (32)				& 541								& 377					& 173.01							& 34.68						& 107.87
			\end{tabular}
		}
	\end{table}
    
    When calculating the cost associated with layers that use MACs, we take account of four accesses to the memory: three for reading activation, weight, and previous partial result, and one for writing the updated partial result. 
    Note that when an activation is a binary value, only a single bit has to be read from the memory. 

\subsection{Results \& Discussion}
\subsubsection{Multi-Layer Perceptron} \label{sec:mlp}
	The first neural network studied in this section is an MLP with three hidden layers and 100 neurons per layer (Net~1.1). 
    The sign non-linearity is applied to all hidden layers as activation function according to Algorithm~\ref{alg:forward} (Net~1.1.a). 
    The second and third hidden layers have binary input and output activations and have been optimized using Algorithm~\ref{alg:optimization} (Net~1.1.b). 
    Each of these layers is considered as a macro-pipeline stage and is not further micro-pipelined for increasing the throughput. 
    The first and last layers of this network are implemented using floating-point MACs (fixed-point quantization is not applied). 
    As a result, there will be no savings in the first layer. 
    However, as explained earlier, there will be about 25\% memory savings in the last layer because the activations are binary values. 

    The classification accuracy and hardware cost of implementing Net~1.1 is compared to a network with the same architecture, but with ReLU activation function. 
    The operations in the latter network are implemented once using single-precision floating-point MACs (Net~1.2) and another time using half-precision floating-point MACs (Net~1.3). 

    Table~\ref{table:mlp-accuracy} compares classification accuracy of the aforementioned networks. 
	It is observed that quantization of activations (Net~1.1.a) has caused a 1.38\% accuracy degradation compared to networks trained with ReLU non-linearity. 
    Moreover, it is observed that optimizing the second and third hidden layers (Net~1.1.b) has increased the classification accuracy by 0.12\% compared to Net~1.1.a. 

    \begin{table}[h]
		\centering
		\captionsetup{justification=centering}
		\caption{Classification accuracy of different MLPs}
		\label{table:mlp-accuracy}
		\resizebox{0.8\columnwidth}{!}{%
			\begin{tabular}{l cccc}
            \textbf{Network}		& Net 1.1.a				& Net 1.1.b			& Net 1.2		& Net 1.3		\\
            \midrule
            \textbf{Accuracy} (\%)	& 96.89					& 97.01				& 98.27			& 98.27
			\end{tabular}
		}
	\end{table}
    
    Table~\ref{table:mlp-hw} presents the hardware cost of the second and third hidden layers of Net~1.1.b on the target FPGA. 
    Implementing these layers in Net~1.2 and Net~1.3 requires 20,000 MAC operations. 
    Obviously, this high number of operations have to be performed on the target platform during several cycles. 
    However, the optimized implementation presented in Table~\ref{table:mlp-hw} calculates all outputs in parallel with a latency that is equal to $0.88 \times$ latency of a \textit{single} 32-bit MAC and $1.44 \times$ latency of a \textit{single} 16-bit MAC. 
    The optimized realization consumes 112,173 ALMs, which is about $207 \times$ that of a 32-bit MAC and $575 \times$ that of a 16-bit MAC. 
    Assuming that all 20,000 MACs in Net~1.2 and Net~1.3 could be realized in parallel, the number of ALMs used in the optimized realization is about $97 \times$ and $35 \times$ lower than the number of ALMs used for Net~1.2 and Net~1.3, respectively. 
    Last but not least, the optimized representation needs to read/write 400 bits from/to memory while 32-bit and 16-bit MAC-based representations need to read/write \SI{312.5}{KB} and \SI{156.25}{KB} from/to memory, respectively. 
    This is equivalent to $6,400 \times$ and $3,200 \times$ savings in accesses to the memory. 
    These substantial savings in memory accesses lead to enormous latency reduction and energy savings according to Table~\ref{table:intel-delay} and Table~\ref{table:energy-45nm}. 

	\begin{table}[h]
		\centering
		\captionsetup{justification=centering}
		\caption{Hardware realization results for second and third hidden layers of Net 1.1.b}
		\label{table:mlp-hw}
		\resizebox{0.9\columnwidth}{!}{%
			\begin{tabular}{ccccc}
				\multirow{2}{*}{\textbf{ALMs}}	& \textbf{Registers}	& \textbf{Clock Frequency}			& \textbf{Latency}			& 								\textbf{Power}				\\
                {}								& (bits)				& (\si{\MHz})						&  (\si{\ns})				& 								(\si{\mW})					\\
				\midrule
                112,173							& 302					& 65.3								& 30.63						&								396.46
			\end{tabular}
		}
	\end{table}

	It is needless to say that the overall savings in the network are smaller than the aforementioned values because of the cost associated with the first and last layers. 
    However, as the networks get deeper, the first and last layers will play a less important role in the overall network performance. 
    Table~\ref{table:mlp-compare} compares the cost of implementing different layers of Net~1.1.b and Net~1.2. 
    To allow comparison of our work with MAC-based implementations, we present the cost associated with the second and third hidden layers in terms of MACs (through dividing the number of ALMs used in these layers by the number of ALMs used in a single MAC). 
    It is observed that Net~1.1.b requires \SI{79.61}{k} MAC operations and has to read/write \SI{1.21}{MB} data from/to memory. 
    On the other hand, Net~1.2 requires \SI{99.4}{k} MAC operations and has to read/write \SI{1.52}{MB} data from/to memory. 
    This translates into 20\% savings in computations and 20\% savings in memory accesses for implementing the whole network. 

    For the other network discussed in this section, we do not explain the layer-by-layer costs in detail and report the savings briefly. 

	\begin{table}[h]
		\centering
		\captionsetup{justification=centering}
		\caption{Cost of realizing different layers of Net~1.1.b and Net~1.2}
		\label{table:mlp-compare}
        
        \subfloat[][Net~1.1.b] {
        \resizebox{0.5\columnwidth}{!}{%
			\begin{tabular}[]{c cc}
				\multirow{2}{*}{\textbf{Layer}}			& \multirow{2}{*}{\textbf{MACs}}				& \textbf{Memory}		\\
                {}										& {}											& (Bytes)				\\
                \midrule
                FC1										& 78,400										& 1,254,400				\\
                FC2 $+$ FC3								& 207											& 50					\\
                FC4										& 1,000											& 12,125				\\
                \midrule
                Total									& 79,607										& 1,266,575
			\end{tabular}
        }
        }
        \subfloat[][Net~1.2] {
        \resizebox{0.44\columnwidth}{!}{%
			\begin{tabular}[]{c cc}
				\multirow{2}{*}{\textbf{Layer}}			& \multirow{2}{*}{\textbf{MACs}}				& \textbf{Memory}		\\
                {}										& {}											& (Bytes)				\\
                \midrule
                FC1										& 78,400										& 1,254,400				\\
                FC2										& 10,000										& 160,000				\\
                FC3										& 10,000										& 160,000				\\
                FC4										& 1,000											& 16,000				\\
                \midrule
                Total									& 99,400										& 1,590,400
			\end{tabular}
        }
        }
	\end{table}

\subsubsection{Convolutional Neural Network (CNN)}

	The second neural network studied in this section is a CNN, where the first convolutional layer implements $3 \times 3$ convolutions and has 10 output channels while the second convolutional layer implements $3 \times 3$ convolutions and has 20 output channels (Net~2.1). 
    Both convolutional layers are followed by $2 \times 2$ max pooling and the sign non-linearity is applied to all hidden layers as activation function according to Algorithm~\ref{alg:forward} (Net~2.1.a). 
    The second convolutional layer has binary input and output activations and has been optimized using Algorithm~\ref{alg:optimization} (Net~2.1.b). 
    Implementation of first and last layers and pipelining method are similar to Net~1.1.b. 

    Similar to the previous section, the cost of implementing Net~2.1 is compared with a network with the same architecture, but with ReLU activation function. 
    Net~2.2 and Net~2.3 represent implementations of the latter network using single-precision and half-precision floating-point operations, respectively. 

	Table~\ref{table:cnn-accuracy} compares classification accuracy of aforementioned networks. 
    It is observed that quantization of activations (Net~2.1.a) has caused a 0.79\% accuracy degradation compared to networks trained with ReLU non-linearity. 
    In addition, it is observed that using ISFs in Net~2.1.b has decreased classification accuracy by 0.29\% compared to the case where outputs are calculated using dot products (Net~2.1.a). 

    \begin{table}[h]
		\centering
		\captionsetup{justification=centering}
		\caption{Classification accuracy of different CNNs}
		\label{table:cnn-accuracy}
		\resizebox{1\columnwidth}{!}{%
			\begin{tabular}{l cccc}
            \textbf{Network}		& Net 2.1.a				& Net 2.1.b			& Net 2.2		& Net 2.3		\\
            \midrule
            \textbf{Accuracy} (\%)	& 98.21					& 97.92				& 99.00			& 99.00
			\end{tabular}
		}
	\end{table}
    
    Table~\ref{table:cnn-hw} presents the hardware cost of realizing the kernels of the second convolutional layer of Net~2.1.b on the target FPGA. 
    Implementing the convolution on each patch of input in Net~2.2 and Net~2.3 requires 1,800 MAC operations. 
    The optimized implementation presented in Table~\ref{table:cnn-hw} calculates all outputs in parallel with a latency that is equal to $0.41 \times$ latency of a \textit{single} 32-bit MAC and $0.67 \times$ latency of a \textit{single} 16-bit MAC. 
    The optimized realization consumes 15,990 ALMs, which is about $30 \times$ that of a 32-bit MAC and $82 \times$ that of a 16-bit MAC. 
    Assuming that all 1,800 MACs in Net~2.2 and Net~2.3 could be realized in parallel, the number of ALMs used for realizing the aforementioned kernels is about $60 \times$ and $22 \times$ lower than the number of ALMs used for Net~2.2 and Net~2.3. 
    Last but not least, the optimized representation needs to read/write 110 bits from/to memory for each patch of the input while 32-bit and 16-bit MAC-based representations need to read/write \SI{28.13}{KB} and \SI{14.06}{KB} from/to memory, respectively. 
    This is equivalent to $2095 \times$ and $1047 \times$ savings in accesses to the memory. 

	\begin{table}[h]
		\centering
		\captionsetup{justification=centering}
		\caption{Hardware realization results for the second convolutional layer of Net 2.1.b}
		\label{table:cnn-hw}
		\resizebox{0.9\columnwidth}{!}{%
			\begin{tabular}{ccccc}
				\multirow{2}{*}{\textbf{ALMs}}	& \textbf{Registers}	& \textbf{Clock Frequency}			& \textbf{Latency}			& 								\textbf{Power}				\\
                {}								& (bits)				& (\si{\MHz})						&  (\si{\ns})				& 								(\si{\mW})					\\
				\midrule
                15,990							& 110					& 70.12								& 14.26						&								41.77
			\end{tabular}
		}
	\end{table}

    In this example, Net~2.1.b requires \SI{69.47}{k} MAC operations and has to read/write \SI{1011.45}{KB} data from/to memory. 
    On the other hand, Net~2.2 requires \SI{283.64}{k} MAC operations and has to read/write \SI{4.33}{MB} data from/to memory. 
    This translates into 76\% savings in computations and 77\% savings in memory accesses for implementing the whole network. 

\balance

\section{Conclusion} \label{sec:conclusion}

	In this paper, we presented a method for efficient realization of neural networks through reformulation of the realization problem into a Boolean logic optimization problem. 
    In this method, layers of a neural network are trained to have binary input and output activations. 
    This allows treating each layer as a multi-input multi-output Boolean function, which can be optimized using Boolean logic optimization algorithms. 
    Our experimental results show substantial savings in memory accesses and resource consumption. 

	\section*{Acknowledgements}	
	Mahdi Nazemi proposed the idea of NullaNet as well as realizations based on technology mapping and input enumeration. 
	He also performed machine learning experiments and implemented different steps of realization. 
	Ghasem Pasandi proposed the idea of performing offline simulation for extracting the truth table of a neuron and subsequently optimizing its Boolean function.

	\bibliographystyle{unsrt}
	{\footnotesize
		\bibliography{NullaNet}
	}
	
\end{document}